\def\year{2021}\relax
\documentclass[letterpaper]{article} 
\usepackage{aaai21}  
\usepackage{times}  
\usepackage{helvet} 
\usepackage{courier}  
\usepackage[hyphens]{url}  
\usepackage{graphicx} 
\usepackage{natbib}  
\usepackage{caption} 
\usepackage{algorithm}
\usepackage{algorithmic}
\usepackage{amsmath}
\usepackage{amsfonts, amssymb}
\usepackage{multirow}
\usepackage{makecell}
\usepackage[switch]{lineno}
\title{Self-Domain Adaptation for Face Anti-Spoofing}
\author{
    Jingjing Wang,
    Jingyi Zhang,
    Ying Bian,
    Youyi Cai,\\
    Chunmao Wang,
    Shiliang Pu \thanks{Shiliang Pu is the Corresponding Author.}
    \\
}
\affiliations{
    Hikvision Research Institute\\
    \{wangjingjing9,zhangjingyi,bianying,caiyouyi,wangchunmao,pushiliang.hri\}@hikvision.com
}

\begin{document}
\maketitle

\begin{abstract}
Although current face anti-spoofing methods achieve promising results under intra-dataset testing, they suffer from poor generalization to unseen attacks. Most existing works adopt domain adaptation (DA) or domain generalization (DG) techniques to address this problem. However, the target domain is often unknown during training which limits the utilization of DA methods. DG methods can conquer this by learning domain invariant features without seeing any target data. However, they fail in utilizing the information of target data. In this paper, we propose a self-domain adaptation framework to leverage the unlabeled test domain data at inference. Specifically, a domain adaptor is designed to adapt the model for test domain. In order to learn a better adaptor, a meta-learning based adaptor learning algorithm is proposed using the data of multiple source domains at the training step. At test time, the adaptor is updated using only the test domain data according to the proposed unsupervised adaptor loss to further improve the performance. Extensive experiments on four public datasets validate the effectiveness of the proposed method.
\end{abstract}

In recent years, face recognition (FR) systems have been widely applied in our daily lives, such as smartphones unlock, access control and pay-with-face. However, easy-accessible human face images and various types of presentation attacks (e.g. photo, video replay, or 3D facial mask) make the FR systems vulnerable to spoofing attacks. Therefore, face anti-spoofing has become a crucial part to guarantee the security of these systems and drawn increasing attention in the face recognition community.

Various face anti-spoofing methods have been proposed, and can be categorized into texture-based methods and temporal-based methods. Texture-based methods utilize various appearance cues, such as color \cite{2017Face}, distortion cues \cite{2015Face}), or deep features \cite{2014Learn} to differentiate real and fake faces. While, temporal-based methods leverage various temporal cues, such as facial motions \cite{2007Real, 2017Deep, 2018Joint} or rPPG \cite{20163D, 2018Remote, 2018Learning}. Although these methods achieve promising results in intra-dataset experiments, the performance dramatically degrades in cross-dataset experiments where training and testing data are from different datasets.

\begin{figure}[t!]
\centering
\includegraphics[width=1\linewidth]{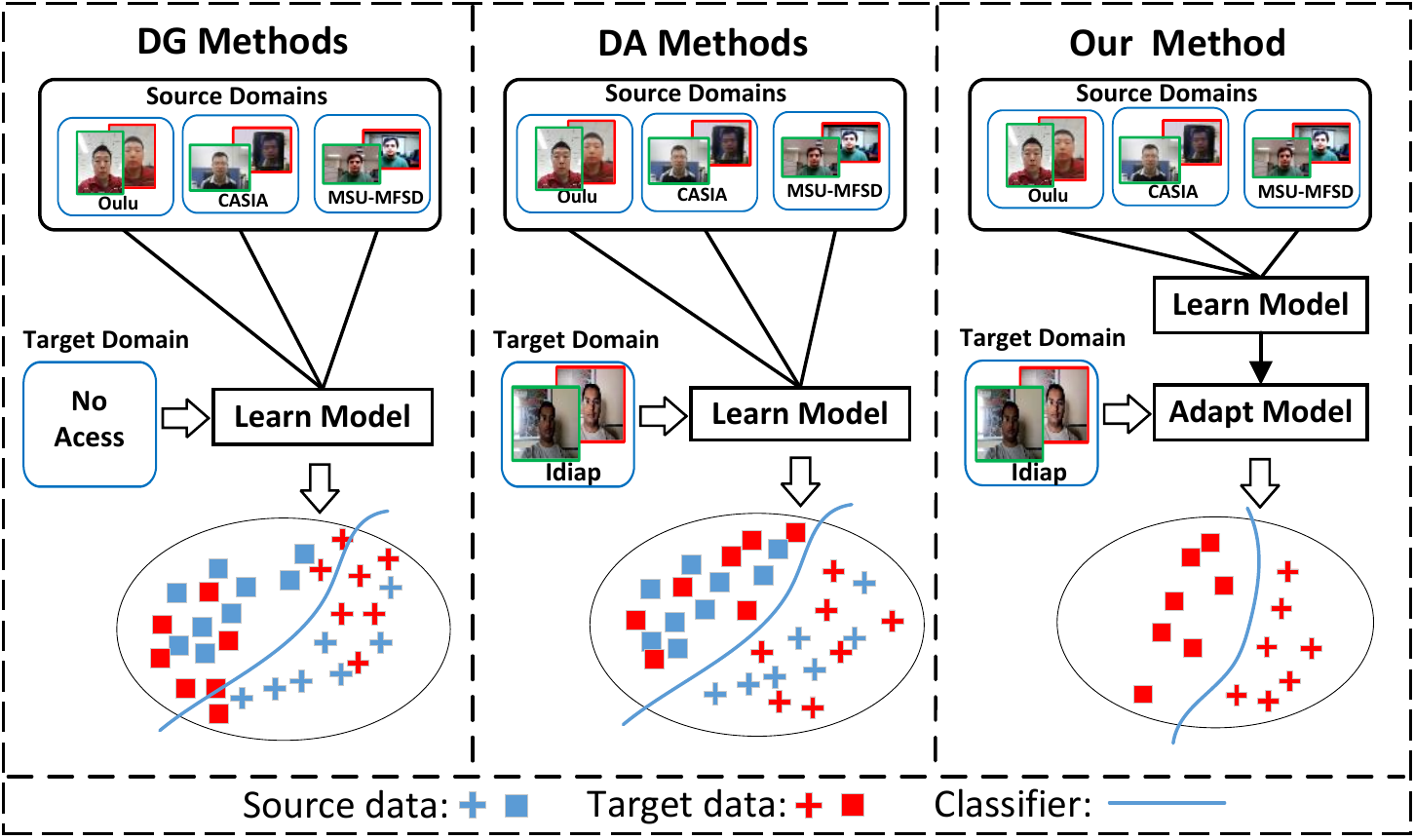}
\caption{Framework comparison among our proposed method and those of DG and DA methods. The DG methods train the model without target domain data, which lose useful information in the target domain. The DA methods need the target domain data to learn the model which is not realistic for face anti-spoofing applications. Our self-domain adaptation method can leverage the target domain information at inference to adapt the model itself for better prediction results on the target domain.}
\label{fig:intro}
\end{figure}

The main reason of the performance drop of previous methods is that the feature distribution between the source and the target domain data is disparate. To make the algorithm generalize well to unseen scenarios, recent face anti-spoofing methods mainly adopt domain adaptation (DA) or domain generalization (DG) techniques to reduce the impact of domain shift. DA based approaches \cite{2018Unsupervised, 2019Improving} adopt DA techniques to minimize the distribution discrepancy between the source and the target domain by leveraging the labeled source domain data and unlabeled target domain data. However, in real scenarios, it is often the case that collecting adequate target domain data is difficult and even no information about the target domain is available during training.

To overcome this limitation of DA based methods, DG based methods \cite{2019Multi, 2020Single, 2020Regularized} address the face anti-spoofing problem in a more realistic scenario assuming no access to target domain information. To this end, multiple source domains are exploited to learn a shared domain agnostic feature space which can generalize well to unseen target domain. However, at test time, when the extracted features of the target domain is mapped to the shared feature space, test domain specific information is lost. We argue that individual domains contain unique characteristics which are discriminative and can aid face anti-spoofing if leveraged appropriately at test time.

\begin{figure}[t!]
\centering
\includegraphics[width=1\linewidth]{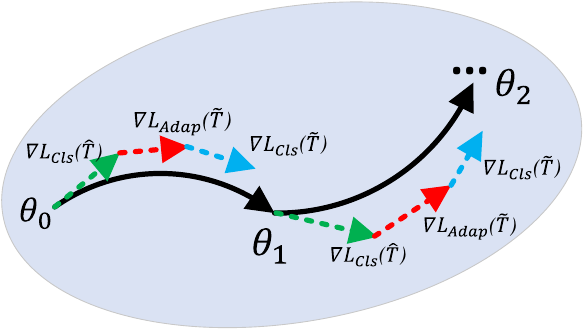}
\caption{Illustration of learning directions of our meta-learning based adaptor learning algorithm. $\nabla L_{Cls(\hat{T})}$, $\nabla L_{Adap(\tilde{T})}$, $\nabla L_{Cls(\tilde{T})}$ denote the directions of supervised classification loss on meta-train domain $\hat{T}$, unsupervised adaptor loss on meta-test domain $\tilde{T}$ and supervised classification loss on meta-test domain $\tilde{T}$ respectively.}
\label{fig:intro2}
\end{figure}

To this end, we propose a self-domain adaptation framework to leverage information of the test domain. The comparison among the proposed framework and those of DA and DG based methods is illustrated in Fig. \ref{fig:intro}. We first learn an adaptor utilizing the labeled data of multiple source domains. Then, at inference the adaptor is optimized to leverage the feature distribution of the test domain. Through the two-step learning, a well-initialized adaptor can be learned for test-time domain adaptation and the test domain doesn't need to be accessible during the training stage. More specifically, in order to learn the adaptor, a meta-learning based adaptor learning algorithm is proposed. After randomly sampling one domain as the meta-train domain and another as the meta-test domain from the source domains, the model is firstly updated according to the classification loss on the meta-train domain. Then it is optimized according to the proposed unsupervised adaptor loss on the meta-test domain and finally updated via the classification loss on the meta-test domain. In this way, the adaptor is optimized towards the direction which is more efficient and discriminative for unsupervised test-time domain adaptation at inference. The learning directions of the meta-learning algorithm are illustrated in Fig. \ref{fig:intro2}.

The main contributions of this work are summarized as follows: 1) In contrast to most state-of-the-art works which suppress the domain-specific information, we are the first to propose a self-domain adaptation framework to leverage the test domain information, which opens up a new direction for the face anti-spoofing community. 2) We propose a meta-learning based adaptor learning algorithm for better adaptor initialization, and an unsupervised adaptor loss for appropriate adaptor optimization. 3) We make comprehensive comparisons and show the promising performance of our self-domain adaptation method on four public datasets.

\section{Related Work}

\textbf{Traditional Face Anti-spoofing Methods.}
Face anti-spoofing methods can be coarsely divided into two groups: texture-based methods and temporal-based methods. Texture-based methods differentiate real and fake faces via different texture cues. Prior researchers mainly extract handcrafted features, e.g. LBP \cite{2017Face}, HoG \cite{2014Context}, SIFT \cite{2016Secure} and SURF \cite{2016Face} and train a binary classifier to discern the live vs. spoof, such as SVM and LDA. Recently, deep learning based face anti-spoofing methods show significant improvement over the conventional ones. More researchers turn to employ CNNs to extract more discriminative features \cite{2014Learn, 2018Face, 2018De-Spoofing}.
On the other hand, temporal-based methods extract different temporal cues in consecutive frames for spoofing face detection. Early works use particular liveness facial motions, such as mouth motion \cite{2007Real} and eye-blinking \cite{2007Eyeblink, 2007Blinking} as the temporal cues. While, recent works learn more general temporal cues. Xu et al. \cite{2016Learning} utilize a CNN-LSTM architecture for effective temporal feature learning. Other researchers \cite{20163D, 2018Remote, 2018Learning} propose to capture discriminative rPPG signals as robust temporal features. Although remarkable results are obtained by above methods in intra-dataset testing, dramatic performance degradation is observed in cross-dataset testing. The main reason is that they discard the relationship among different domains and extract dataset-biased features.

\textbf{DA \& DG based Face Anti-spoofing Methods.}
To make the algorithm generalize well to unseen scenarios, most recent face anti-spoofing methods turn to adopt DA and DG techniques to learn domain invariant features. Li et al. \cite{2018Unsupervised} and Wang et al. \cite{2019Improving} make the learned feature space domain invariant by minimizing MMD \cite{gretton2012a} and adversarial training respectively. Our work is also related to some unsupervised domain adaptation methods \cite{2018Semantic, 2018CoTraining} based on self-training, since we both using unsupervised learning on test domain. However, these methods assume access to the target domain data during training, which is not realistic in most situations. To overcome this limitation, DG techniques are exploited to extract domain invariant features without target domain data. Shao et al. \cite{2019Multi} propose to learn a generalized feature space via a novel multi-adversarial discriminative deep domain generalization framework. Jia et al. \cite{2020Single} propose a single-side domain generalization framework for face anti-spoofing considering that only the real faces from different domains should be undistinguishable, but not for the fake ones. The most related work to ours is proposed by Shao et al. \cite{2020Regularized}, where a generalized feature space is learned by a new fine-grained meta-learning strategy. However, they utilize meta-learning to learn domain invariant features without leveraging the target domain information, while we employ meta-learning to learn a domain adaptor which can update itself using the target domain data at inference.

\textbf{Self-domain adaptation methods.}
Self-domain adaptation adapts the deployed model to various target domains during inference without accessing the source data which is very suitable for the situation of face anti-spoofing. Qin et al. \cite{2020one} propose a one-class domain adaptation face anti-spoofing method without source domain data. However they need living faces for adaptation on the test domain. How to adapt the model itself to the test domain unsupervisedly, has received less attention. Li et al. \cite{2020Model} propose a novel source-free unsupervised domain adaptation framework through generating labeled target-style data by a conditional generative adversarial net. Wang et al. \cite{2020Fully} adapt the model by modulating its representation with affine transformations to minimize entropy. He et al. \cite{2020Self} propose to transform the input test image and features via adaptors to reduce the domain shift measured by auto-encoders. These works inspire us to propose a self-domain adaptation framework for face anti-spoofing.

\begin{figure*}[ht!]
\centering
\includegraphics[width=1\linewidth]{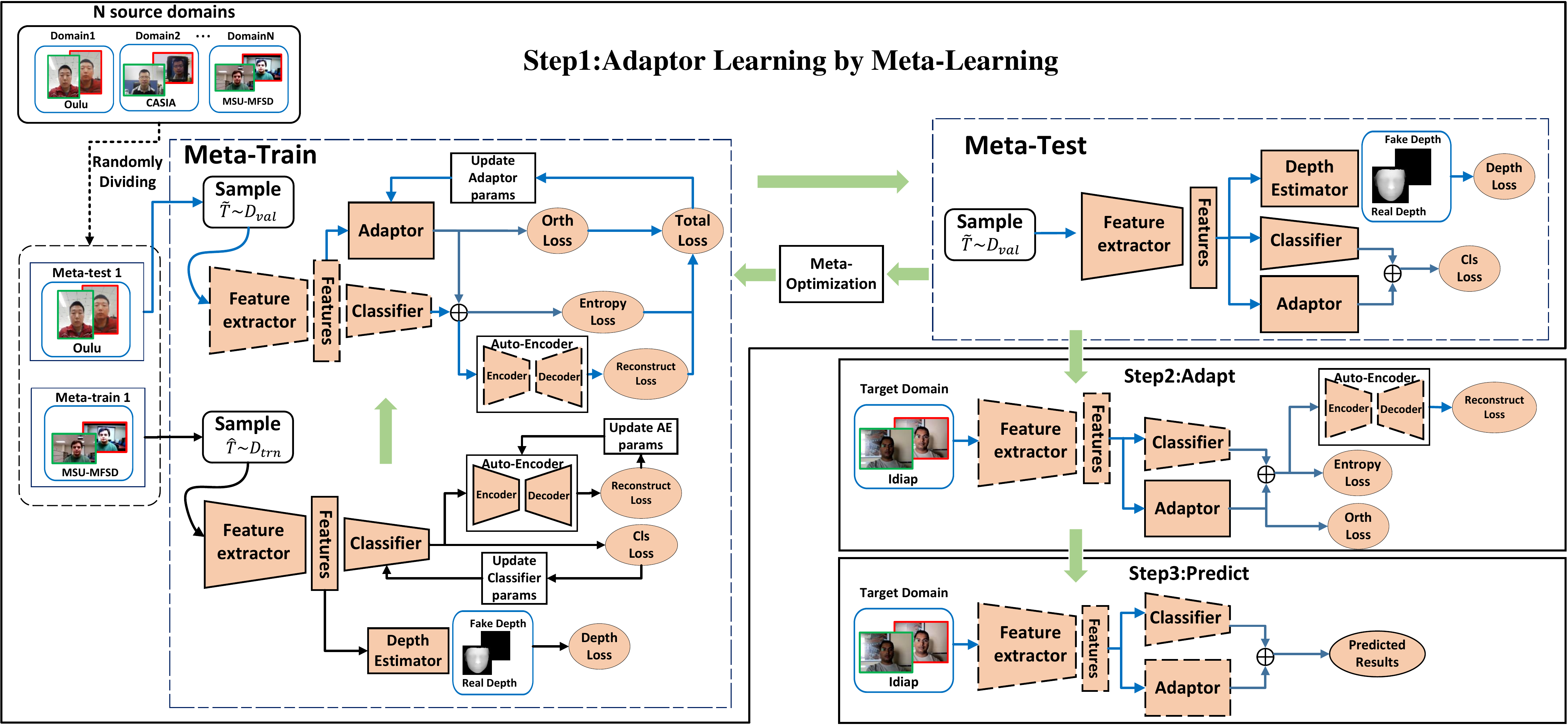}
\caption{Overview of our self-domain adaptation framework. The module with solid lines means it is being trained while the one with dashed lines indicates that its parameters are fixed.}
\label{fig:overview}
\end{figure*}

\section{Proposed Method}
\subsection{Overview}
To leverage the distribution of target data at inference, we propose to learn an adaptor using the data of multiple available source domains at training step and adjust the adaptor with only unlabeled test domain data at inference. Our method is divided into three main procedures. Firstly, in addition to the discriminative feature learning using the available labeled source domains, an adaptor is learned simultaneously. Secondly, at inference, all the other parameters are fixed and only the adaptor is updated according to the real test domain data unsupervisedly. Finally, the model is fixed, and testing is run based on the fixed model. The framework of our method is illustrated in Fig. \ref{fig:overview}.

\subsection{Learning Adaptor at Training}
\begin{figure}[t]
\centering
\includegraphics[width=1\linewidth]{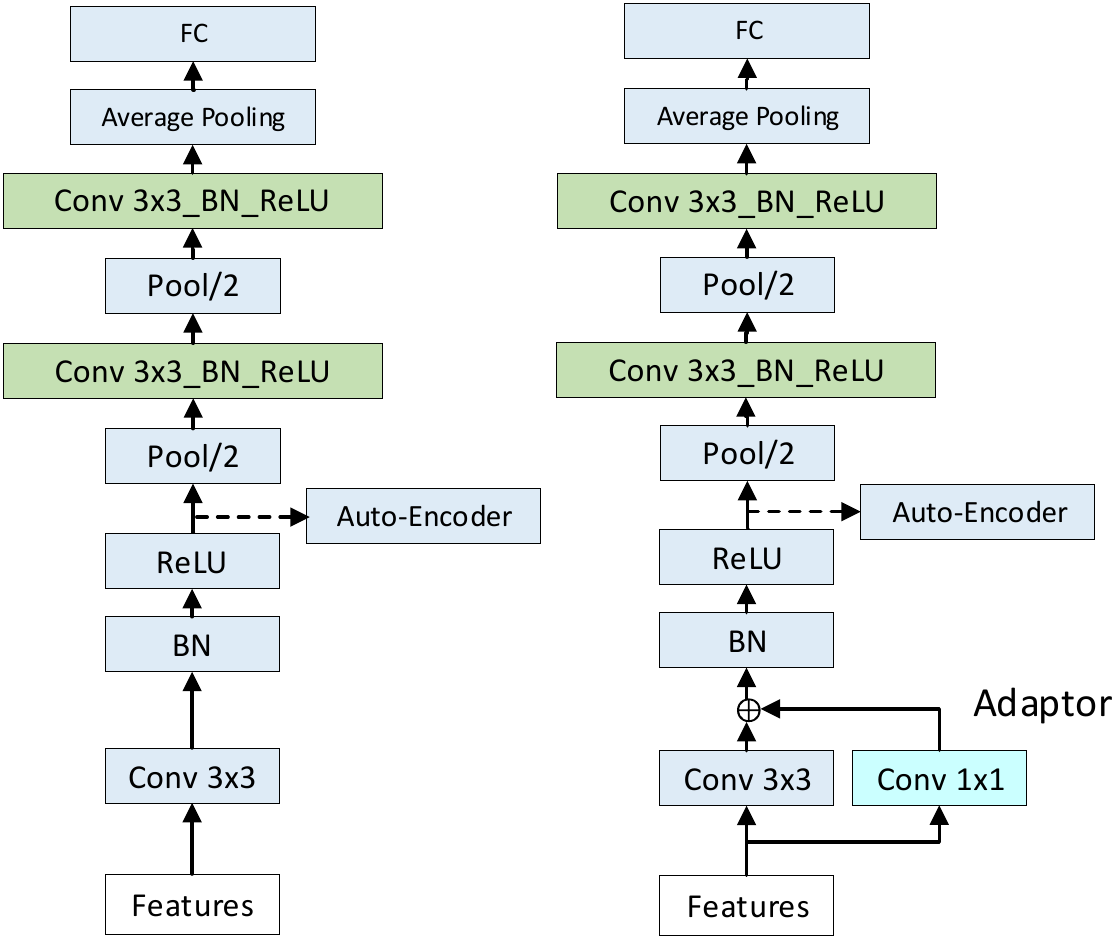}
\caption{The network structure of our classifier with adaptor (right) and without adaptor (left).}
\label{fig:structure}
\end{figure}

We propose a meta-learning based adaptor learning algorithm to obtain a better initialized adaptor which can adapt to the test domain efficiently at inference. The network proposed in our framework during training is composed of a feature extractor (denoted as $F$), a classification header (denoted as $C$), a depth estimator (denoted as $D$), an adaptor (denoted as $A$), and an Autoencoder (denoted as $R$) as shown in Fig. \ref{fig:overview}. The detailed structure of $C$ and the connections among $C$, $A$ and $R$ are illustrated in Fig. \ref{fig:structure}. We think the test domain has its own domain specific information which can be learned by domain specific kernels and can be added into the common features. For efficiency, the adaptor is designed as a $1\times1$ convolution and connected to the first convolution of $C$ through a residual architecture. Another benefit is that after removing the adaptor, the model can return to the original one. We denote the classifier with adaptor as $C_a$. The feature maps after the first activation layer of $C$ and $C_a$ are used for reconstruction by $R$. We denote the first layers (up to the first activation layer) of $C$ and $C_a$ as $C_{l_1}$ and $C_{a,l_1}$. Suppose that we have access to $N$ source domains of face anti-spoofing task, denoted as $\mathbb{D} = [D_1, D_2, \cdots, D_N]$. In order to simulate the real scenarios, one source domain is randomly chosen from $\mathbb{D}$ as the meta-train domain $D_{trn}$, and another different domain is selected as the meta-test domain $D_{val}$. The adaptor learning algorithm composes of two main steps: meta-train and meta-test. The whole meta-learning process is illustrated in the Step1 of Fig. \ref{fig:overview} and summarized in Algorithm \ref{alg:Meta-Learning}.

\textbf{Meta-Train.}
During meta-train, we have access to the labeled source domain $D_{trn}$ and unlabeled target domain $D_{val}$. We sample batches in $D_{trn}$ as $\hat{T}$ and batches in $D_{val}$ as $\tilde{T}$. The cross-entropy classification loss is conducted based on the binary class labels in $\hat{T}$ as follows:
\begin{flalign}
&\mathcal{L}_{Cls(\hat{T})}(\theta_F, \theta_C) =  \notag \\
& -\sum\limits_{(x,y)\sim\hat{T}}y log C(F(x)) + (1-y)log(1-C(F(x)))
\end{flalign}
where $\theta_F$ and $\theta_C$ are the parameters of the feature extractor and the classifier. The updated classifier $C^\prime$ can be calculated as ${\theta_C}^\prime = \theta_{C} - \alpha \nabla_{\theta_C}\mathcal{L}_{Cls(\hat{T})}(\theta_F, \theta_C)$. Meanwhile, as suggested by \cite{2018Learning, 2019Multi, 2020Regularized}, we also incorporate face depth information as auxiliary supervision in the learning process as follows:
\begin{equation}
\mathcal{L}_{Dep(\hat{T})}(\theta_F, \theta_D) = \sum\nolimits_{(x,I_d)\sim\hat{T}}||D(F(x))-I_d||^2
\end{equation}
where $\theta_D$ is the parameter of the depth estimator and $I_d$ are
the pre-calculated face depth maps of input face images. In addition, we train an autoencoder to reconstruct the feature maps of $C$ as follows:
\begin{equation}
\mathcal{L}_{AE(\hat{T})}(\theta_{R}; \theta_F, {\theta_C})= \sum\limits_{x\sim\hat{T}}||R(C_{l_1}(F(x)))- C_{l_1}(F(x))||^2
\end{equation}
where $\theta_R$ is the parameter of the autoencoder. The updated autoencoder $R^\prime$ is calculated as ${\theta_{R}}^\prime = \theta_{R} - \alpha \nabla_{\theta_R} \mathcal{L}_{AE(\hat{T})}(\theta_R; \theta_F, {\theta_C})$.

After the supervised update on the source domain $D_{trn}$, the adaptor $A$ is added into the classifier $C$ to adapt the features of $C$ for the target domain $D_{val}$. Considering that the label of the target domain data is unavailable, we must learn the adaptor unsupervisedly. We take following factors into account. Firstly the feature distribution of the target domain should be similar with the one of the source domain, since the discriminative classifier is mainly learned based on the labeled source domain data. To this end, we utilize the autoencoder trained on the source domain data as the similarity measure, and minimize the reconstruction error on the target domain data as follows:
\begin{equation}
\label{eq:ae}
\mathcal{L}_{AE(\tilde{T})}({\theta_A};\theta_F, {\theta_C}^\prime, {\theta_R}^\prime)= \sum\nolimits_{x\sim\tilde{T}}||R^\prime(F_a)- F_a||^2
\end{equation}
where $\theta_A$ is the parameter of the adaptor, and $F_a = C_{a,l_1}(F(x))$ with ${\theta_C}^\prime$ and $\theta_A$. Secondly, the adaptor should leverage the discriminative information on the target domain to predict more confident scores. Therefore we minimize the entropy of the predicted probability on the target domain, as defined by:
\begin{equation}
\label{eq:Ent}
\mathcal{L}_{Ent(\tilde{T})}({\theta_A};\theta_F, {\theta_C}^\prime, {\theta_R}^\prime)= \sum\limits_{x\sim\tilde{T}}- C_{a}(F(x)) log(C_{a}(F(x)))
\end{equation}
Finally, to prevent the feature mode collapse, we impose orthogonality on $\theta_A$ by using the Spectral Restricted Isometry Property Regularization \cite{bansal2018can}:
\begin{equation}
\label{eq:Orth}
\mathcal{L}_{Orth(\tilde{T})}({\theta_A})= \sigma(\theta_A^T\theta_A-I)
\end{equation}
where $\sigma(\cdot)$ calculates the spectral norm of $\theta_A^T\theta_A-I$. The final unsupervised loss on the adaptor is composed of Equation \ref{eq:ae}, \ref{eq:Ent} and \ref{eq:Orth}:
\begin{equation}
\label{eq:adap_loss}
\mathcal{L}_{Adap(\tilde{T})}({\theta_A}; \theta_F, {\theta_C}^\prime, {\theta_R}^\prime)= \mathcal{L}_{Ent} +\lambda\mathcal{L}_{AE} + \mu \mathcal{L}_{Orth}
\end{equation}
where $\mu$ and $\lambda$ balance the three losses. Based on the unlabeled target data, the updated adaptor $A^\prime$ is calculated as ${\theta_{A}}^\prime = \theta_{A} - \alpha \nabla_{\theta_A} \mathcal{L}_{Adap(\tilde{T})}(\theta_A;\theta_F, {\theta_C}^\prime, {\theta_{R}}^\prime)$.

\textbf{Meta-Test.} By adopting meta-learning based adaptor learning, we encourage our face
anti-spoofing model which is trained on the source domain supervisedly and further adapted on the target domain unsupervisedly can perform well on the target domain. Thus, after self-domain adaptation, the model should have good classification performance on the target domain by minimize the cross-entropy loss defined over the updated ${\theta_A}^\prime$ and ${\theta_C}^\prime$ as:
\begin{flalign}
&\mathcal{L}_{Cls(\tilde{T})}(\theta_F, {\theta_C}^\prime, {\theta_A}^\prime) = \notag \\ &-\sum\nolimits_{(x,y)\sim\tilde{T}}y log C^\prime_a(F(x))+(1-y)log(1-C^\prime_a(F(x)))
\end{flalign}
The face depth information is also incorporated:
\begin{equation}
\mathcal{L}_{Dep(\tilde{T})}(\theta_F, \theta_D) = \sum\nolimits_{(x,I_d)\sim\tilde{T}}||D(F(x))-I_d||^2
\end{equation}

\begin{algorithm}[t!]
	\caption{Adaptor Learning by Meta-Learning}
    \label{alg:Meta-Learning}
	\begin{algorithmic}[1]
        \REQUIRE  \ \\
        \textbf{Input:} N source domains $\mathbb{D} = [D_1, D_2, \cdots, D_N]$ \\
        \textbf{Initialization:} Model parameters $\theta_F$, $\theta_C$ $\theta_D$, $\theta_R$, $\theta_A$. Hyperparameters $\alpha$, $\beta$, $\lambda$, $\mu$.
		\WHILE{not done}
            \STATE Randomly select a domain in $D$ as $D_{trn}$, and another domain in $D$ as $D_{val}$
            \STATE Sampling batch in $D_{val}$ as $\tilde{T}$
            \STATE \textbf{Meta-train:}  Sampling batch in $D_{trn}$ as $\hat{T}$
            \STATE$\mathcal{L}_{Cls(\hat{T})}(\theta_F, \theta_C) = -\sum\nolimits_{(x,y)\sim\hat{T}}y log C(F(x))+(1-y)log(1-C(F(x)))$
            \STATE ${\theta_C}^\prime = \theta_{C} - \alpha \nabla_{\theta_C} \mathcal{L}_{Cls(\hat{T})}(\theta_F, \theta_C)$
            \STATE $\mathcal{L}_{Dep(\hat{T})}(\theta_F, \theta_D) = \sum\nolimits_{(x,I_d)\sim\hat{T}}||D(F(x))-I_d||^2$
            \STATE $\mathcal{L}_{AE(\hat{T})}(\theta_{R}; \theta_F, {\theta_C})= \sum\nolimits_{x\sim\hat{T}}||R(C_{l_1}(F(x)))- C_{l_1}(F(x))||^2$
            \STATE ${\theta_{R}}^\prime = \theta_{R} - \alpha \nabla_{\theta_R} \mathcal{L}_{AE(\hat{T})}(\theta_R; \theta_F, {\theta_C})$

            \STATE $\mathcal{L}_{Adap(\tilde{T})}(\theta_A;\theta_F, {\theta_C}^\prime, {\theta_{R}}^\prime) =\mathcal{L}_{Ent} +\lambda\mathcal{L}_{AE} + \mu \mathcal{L}_{Orth}$
            \STATE ${\theta_{A}}^\prime = \theta_{A} - \alpha \nabla_{\theta_A} \mathcal{L}_{Adap(\tilde{T})}(\theta_A;\theta_F, {\theta_C}^\prime, {\theta_{R}}^\prime)$
            \STATE \textbf{Meta-test:}
            \STATE$\mathcal{L}_{Cls(\tilde{T})}(\theta_F, {\theta_C}^\prime, {\theta_A}^\prime) = -\sum\limits_{(x,y)\sim\tilde{T}}y log C^\prime_a(F(x)) + (1-y)log(1-C^\prime_a(F(x)))$
            \STATE $\mathcal{L}_{Dep(\tilde{T})}(\theta_F, \theta_D) = \sum\nolimits_{(x,I_d)\sim\tilde{T}}||D(F(x))-I_d||^2$
            \STATE \textbf{Meta-optimization:}
            \STATE $\theta_C \leftarrow \theta_C  - \beta \nabla_{\theta_C}(\mathcal{L}_{Cls(\hat{T})}(\theta_F, \theta_C) + \mathcal{L}_{Cls(\tilde{T})}(\theta_F, {\theta_C}^\prime, {\theta_A}^\prime))$
            \STATE $\theta_F \leftarrow \theta_F  - \beta \nabla_{\theta_F}(\mathcal{L}_{Dep(\hat{T})}(\theta_F, \theta_D) + \mathcal{L}_{Cls(\hat{T})}(\theta_F, \theta_C) + \mathcal{L}_{Dep(\tilde{T})}(\theta_F, \theta_D) + \mathcal{L}_{Cls(\tilde{T})}(\theta_F, {\theta_C}^\prime, {\theta_A}^\prime))$
            \STATE $\theta_D \leftarrow \theta_D  - \beta \nabla_{\theta_D}( \mathcal{L}_{Dep(\hat{T})}(\theta_F, \theta_D) + \mathcal{L}_{Dep(\tilde{T})}(\theta_F, \theta_D))$
            \STATE $\theta_A \leftarrow \theta_A  - \beta \nabla_{\theta_A}(\mathcal{L}_{Adap(\tilde{T})}(\theta_A;\theta_F, {\theta_C}^\prime, {\theta_{R}}^\prime) + \mathcal{L}_{Cls(\tilde{T})}(\theta_F, {\theta_C}^\prime, {\theta_A}^\prime) $
            \STATE $\theta_R \leftarrow {\theta_R}^\prime$
        \ENDWHILE

        \textbf{Return:} Model parameters $\theta_F$, $\theta_C$ $\theta_D$, $\theta_R$, $\theta_A$.\\
	\end{algorithmic}
\end{algorithm}

\subsection{Adapting at Inference}
Given the well-initialized adaptor, during testing, we first optimize the adaptor using only the unlabeled test domain data with all the other parameters fixed using Equation \ref{eq:adap_loss}. Once the update procedure is completed, we keep all the parameters of the model fixed, and test the model on the test domain data. It is noted that during this procedure we have no access to the source domain
data.

\section{Experiments}
\subsection{Experimental Settings}
\textbf{Datasets.}
Four public face anti-spoofing datasets are utilized to evaluate the effectiveness of our method: OULU-NPU \cite{2017OULU} (denoted as O), CASIA-MFSD \cite{Zhang2012A} (denoted as C), Idiap Replay-Attack \cite{2012Replay} (denoted as I), and MSU-MFSD \cite{2015Face} (denoted as M). Following the setting in \cite{2019Multi}, one dataset is treated as one domain in our experiment. We randomly select three datasets as source domains for training and the remaining one as the target domain for testing. Thus, we have four testing tasks in total: O\&C\&I to M, O\&M\&I to C, O\&C\&M to I, and I\&C\&M to O. Following the work of \cite{2019Multi}, the Half Total Error Rate (HTER) and the Area Under Curve (AUC) are used as the evaluation metrics.

\textbf{Implementation Details.}
The detailed structure of the proposed network is illustrated in Tab. 1 in the supplementary material. The size of face image is 256$\times$256$\times$6, where we extract RGB and HSV channels of each face image. The Adam optimizer \cite{2014Adam} is used for the optimization. The learning rates $\alpha$, $\beta$ are set as 1e-3. $\mu$, $\lambda$ are set to 10, 0.1 respectively. During training, the batch size is 20 per domain, and thus 40 for training domains totally. At inference, for efficiency we only optimize the adaptor for one epoch and the batch size is set to 20.
\subsection{Experimental Comparison}

\begin{figure*}[t!]
\centering
\includegraphics[width=1\linewidth]{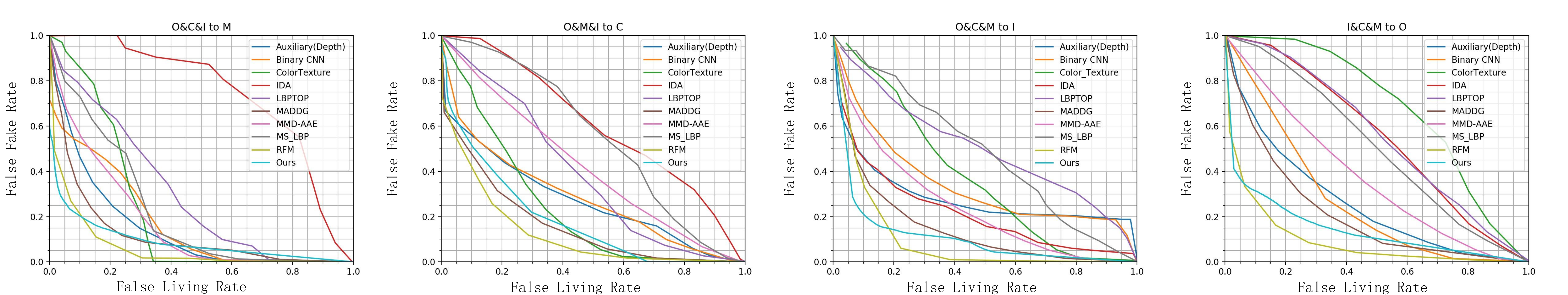}
\caption{ROC curves of four testing tasks for domain adaptation and generation on face anti-spoofing.}
\label{fig:ROC}
\end{figure*}

\begin{table*}[t!]
\centering
\begin{center}
\begin{tabular}{c | c c | c c | c c | c c }
\hline
\multirow{2}{*}{\textbf{Methods}} &
\multicolumn{2}{c|}{\textbf{O\&C\&I to M}} &
\multicolumn{2}{c|}{\textbf{O\&M\&I to C}} &
\multicolumn{2}{c|}{\textbf{O\&C\&M to I}} &
\multicolumn{2}{c}{\textbf{I\&C\&M to O}}\\

    &HTER(\%) &AUC(\%) &HTER(\%) &AUC(\%) &HTER(\%) &AUC(\%) &HTER(\%) &AUC(\%)\\
\hline

IDA &$66.6$ &$27.8$ &$55.1$ &$39.0$ &$28.3$ &$78.2$ &$54.2$ &$44.6$\\

LBPTOP &$36.9$ &$70.8$ &$42.6$ &$61.5$ &$49.5$ &$49.5$ &$53.1$ &$44.0$\\

MS\_LBP &$29.7$ &$78.5$ &$54.2$ &$44.9$ &$50.3$ &$51.6$ &$50.2$ &$49.3$\\

ColorTexture &$28.0$ &$78.4$ &$30.5$ &$76.8$ &$40.4$ &$62.7$ &$63.5$ &$32.7$\\

Binary CNN &$29.2$ &$82.8$ &$34.8$ &$71.9$ &$34.4$ &$65.8$ &$29.6$ &$77.5$\\

Auxiliary(ALL) &- &- &$28.4$ &- &$27.6$ &- &- &-\\

Auxiliary(Depth) &$22.7$ &$85.8$ &$33.5$ &$73.1$ &$29.1$ &$71.6$ &$30.1$ &$77.6$\\

MMD-AAE &$27.0$ &$83.1$ &$44.5$ &$58.2$ &$31.5$ &$75.1$ &$40.9$ &$63.0$\\

MADDG &$17.6$ &$88.0$ &\underline{24.5} &\underline{84.5} &$22.1$ &$84.9$ &$27.9$ &$80.0$\\

RFM &\textbf{13.8} &\textbf{93.9} &\textbf{20.2} &\textbf{88.1} &\underline{17.3} &\textbf{90.4} &\textbf{16.4} &\textbf{91.1}\\
\hline
\textbf{Ours} &\underline{15.4} &\underline{91.8} &\underline{24.5} &$84.4$ &\textbf{15.6} &\underline{90.1} &\underline{23.1} &\underline{84.3}\\
\hline
\end{tabular}
\end{center}
\caption{Comparison to the-state-of-art face anti-spoofing methods on four testing domains. The bold type indicates the best performance, the under-line type indicates the second best performance (the same below).}
\label{tab:SOTA}
\end{table*}

\begin{table*}[t!]
\centering
\begin{center}
\begin{tabular}{c | c c | c c | c c | c c }
\hline
\multirow{2}{*}{\textbf{Methods}} &
\multicolumn{2}{c|}{\textbf{O\&C\&I to M}} &
\multicolumn{2}{c|}{\textbf{O\&M\&I to C}} &
\multicolumn{2}{c|}{\textbf{O\&C\&M to I}} &
\multicolumn{2}{c}{\textbf{I\&C\&M to O}}\\

    &HTER(\%) &AUC(\%) &HTER(\%) &AUC(\%) &HTER(\%) &AUC(\%) &HTER(\%) &AUC(\%)\\
\hline

AdapBN &$20.5$ &{88.0} &$34.5$ &$72.0$ &$27.7$ &$80.3$ &{28.2} &{80.8}\\

FTTA &{20.1} &{88.0} &$35.0$ &$71.2$ &$27.2$ &$79.6$ &$28.3$ &$80.7$\\

SDAN &$17.7$ &$90.0$ &{25.9} &{81.3} &{28.2} &{84.2} &$32.9$ &$75.0$\\

\hline
Ours &\textbf{15.4} &\textbf{91.8} &\textbf{24.5} &\textbf{84.4} &\textbf{16.4} &\textbf{92.0} &\textbf{23.1} &\textbf{84.3}\\
\hline
\end{tabular}
\end{center}
\caption{Comparison to the related self-domain adaptation methods on four testing sets.}
\label{tab:self}
\end{table*}

\textbf{Compared Methods.}
We compare several state-of-the-art face anti-spoofing methods as follows: \textbf{Multi-Scale LBP
(MS\_LBP)} \cite{maatta2011face}; \textbf{Binary CNN} \cite{2014Learn}; \textbf{Image Distortion
Analysis (IDA)}\cite{2015Face}; \textbf{Color Texture (CT)} \cite{2017Face}; \textbf{LBPTOP} \cite{2014dynamic}; \textbf{Auxiliary} \cite{2018Learning}: This method learns a CNN-RNN model to estimate the face depth from one frame and rPPG signals through multiple frames (denoted as Auxiliary(All)). To fairly compare our method only using one frame information, we also compare the results of its face depth estimation component (denoted as Auxiliary(Depth Only)); \textbf{MMD-AAE} \cite{2018Domain}; \textbf{MADDG} \cite{2019Multi}; and \textbf{RFM} \cite{2020Regularized}. Moreover, we also compare the related self-domain adaptation methods in the face anti-spoofing task: \textbf{Adaptive Batch Normalization (AdapBN)} \cite{Li2018Adaptive}; \textbf{Fully Test-Time Adaptation (FTTA)} \cite{2020Fully}; and \textbf{Self Domain Adapted Network (SDAN)} \cite{2020Self}.

\textbf{Comparison Results with SOTA Face Anti-Spoofing Methods.}
From comparison results in Tab. \ref{tab:SOTA} and Fig. \ref{fig:ROC}, it can be seen that the proposed method outperforms most of the state-of-the-art face anti-spoofing methods. This is because all other face anti-spoofing methods except for the three DG methods (MMD-AAE, MADDG, RFM) pay no attention to the intrinsic distribution relationship among different domains and extract dataset-biased features which causes significant performance degradation in case of cross-dataset testing scenarios. Although the MMD-AAE and MADDG exploit the DG techniques to extract domain invariant discriminative cues, they fail in utilizing the distribution of target data which contains domain specific discriminative information for the target domain. On the contrary, our proposed self-domain adaptation methods can leverage the distribution of target data to learn more discriminative features which are specific for the target domain. The only exception is that we achieve slightly worse results compared with RFM. Although our method and RFM both use the meta-learning technique, RFM utilizes it to learn domain invariant features, while we utilize meta-learning to learn a domain adaptor which can adapt to the target domain efficiently. We believe that the two approaches are complementary, and combination of them can further improve the performance. We leave it in the future work.

\textbf{Comparison Results with Related Self-Domain Adaptation Methods.}
The results of Tab. \ref{tab:self} show that our method outperforms all the related self-domain adaptation methods. AdapBN and FTTA only adjust the parameters of BN for target domain adaptation. The overall performances of AdapBN and FTTA are worse than SDAN which verifies only adjusting BN is not sufficient for adequate target domain adaptation for the task of face anti-spoofing. Although SDAN uses a more complex adaptor, it directly adapts the model according to the target domain without learning a well-initialized adaptor in advance. While we leverage multiple source domains to learn an adaptor which can adapt to the target domain efficiently through meta-learning.

\subsection{Ablation Study}
Our method consists of three steps, and the unsupervised adaptor loss consists of three parts. In this subsection, we evaluate the influences of different steps and different parts.

\textbf{Influences of each part of the unsupervised adaptor loss.}
Ours\_wo/$\mathcal{L}_{AE}$, Ours\_wo/$\mathcal{L}_{Orth}$ and Ours\_wo/$\mathcal{L}_{Ent}$ denote the proposed unsupervised adaptor loss without the reconstruction part, the orthogonality part and the entropy part respectively during the adaptor learning and adapting. The results of Tab. \ref{tab:ablation} show that the performances of Ours\_wo/$\mathcal{L}_{AE}$, Ours\_wo/$\mathcal{L}_{Orth}$ and Ours\_wo/$\mathcal{L}_{Ent}$ decrease in different degrees which validates the effectiveness of each part of the proposed adaptor loss.
\begin{table*}[t!]
\centering
\begin{center}
\begin{tabular}{c | c c | c c | c c | c c }
\hline
\multirow{2}{*}{\textbf{Methods}} &
\multicolumn{2}{c|}{\textbf{O\&C\&I to M}} &
\multicolumn{2}{c|}{\textbf{O\&M\&I to C}} &
\multicolumn{2}{c|}{\textbf{O\&C\&M to I}} &
\multicolumn{2}{c}{\textbf{I\&C\&M to O}}\\

    &HTER(\%) &AUC(\%) &HTER(\%) &AUC(\%) &HTER(\%) &AUC(\%) &HTER(\%) &AUC(\%)\\
\hline
Baseline  &$22.7$ &$85.8$ &$33.5$ &$73.1$ &$29.1$ &$71.6$ &$30.1$ &$77.6$\\

Ours\_wo/$\mathcal{L}_{AE}$ &$18.8$ &$87.9$ &$34.5$ &$72.4$ &$27.8$ &$80.0$ &$28.1$ &$80.8$\\

Ours\_wo/$\mathcal{L}_{Orth}$ &{17.6} &{90.4} &$29.9$ &$78.3$ &$29.2$ &$78.6$ &$30.5$ &$81.3$\\

Ours\_wo/$\mathcal{L}_{Ent}$ &$23.0$ &$90.2$ &$26.2$ &$81.6$ &$26.2$ &$84.8$ &$30.3$ &$81.2$\\

Ours\_wo/meta &$23.5$ &$85.5$ &$26.4$ &\textbf{85.1} &$18.2$ &$88.6$ &$30.5$ &$79.5$ \\

Ours\_wo/adapt &$19.4$ &$89.2$ &$32.9$ &$72.2$ &$16.9$ &$89.6$ &$24.4$ &$83.0$\\
\hline
Ours &\textbf{15.4} &\textbf{91.8} &\textbf{24.5} &{84.4} &\textbf{16.4} &\textbf{92.0} &\textbf{23.1} &\textbf{84.3}\\
\hline
\end{tabular}
\end{center}
\caption{Evaluation of different loss parts and different steps of the proposed framework on four testing sets. }
\label{tab:ablation}
\end{table*}

\textbf{Influences of each step of our method.}
Our method consists of three main steps: adaptor learning with meta learning, adaptor optimizing at inference and final testing. Ours denotes the proposed method with all the three steps. Ours\_wo/meta denotes our method without the first step: adaptor learning with meta learning, which initializes the parameters of the adaptor randomly and directly optimizes the adaptor at inference using the proposed adaptor loss. Ours\_wo/adapt denotes our methods without the second step: adaptor optimizing, which neglects the information of the test domain and directly predicts results on the test domain using the adaptor learned by the first step. Baseline denotes learning the model using the source domain data without the adaptor and directly predicts results on the test domain. Tab. \ref{tab:ablation} shows that the proposed method has degraded performance if any step is excluded. Specifically, the results of Ours\_wo/meta verify that the pre-learned adaptor through meta-learning during training benefits the adaptation at inference. The results of Ours\_wo/adapt verify that further optimizing the adaptor to leverage the distribution of the test domain is important to further improve the performance. It is also worth noting that on some test domains using the pre-learned adaptor by meta-learning improves the baseline by a large margin, and the improvement of further optimizing the adaptor on the test domain is marginal, while on some test domains using the pre-learned adaptor by meta-learning improves the baseline marginally, and the improvement of further optimizing the adaptor is significant. For examples, on the O\&C\&M to I set, compared to Baseline, $12.2\%$ HTER improvement is achieved by Ours\_wo/adapt, and only $0.5\%$ HTER improvement is further achieved by Ours, while on the O\&M\&I to C set, compared to Baseline, only $0.5\%$ HTER improvement is achieved by Ours\_wo/adapt and $8.4\%$ HTER improvement is further achieved by Ours. This is because that O\&C\&M to I has little domain shift, so the adaptor learned by meta-learning is already suitable for I, while O\&M\&I to C has significant domain shift, so optimizing the adaptor to leverage the distribution of the test domain can boost the performance significantly.

\begin{figure}[t!]
\centering
\includegraphics[width=1\linewidth]{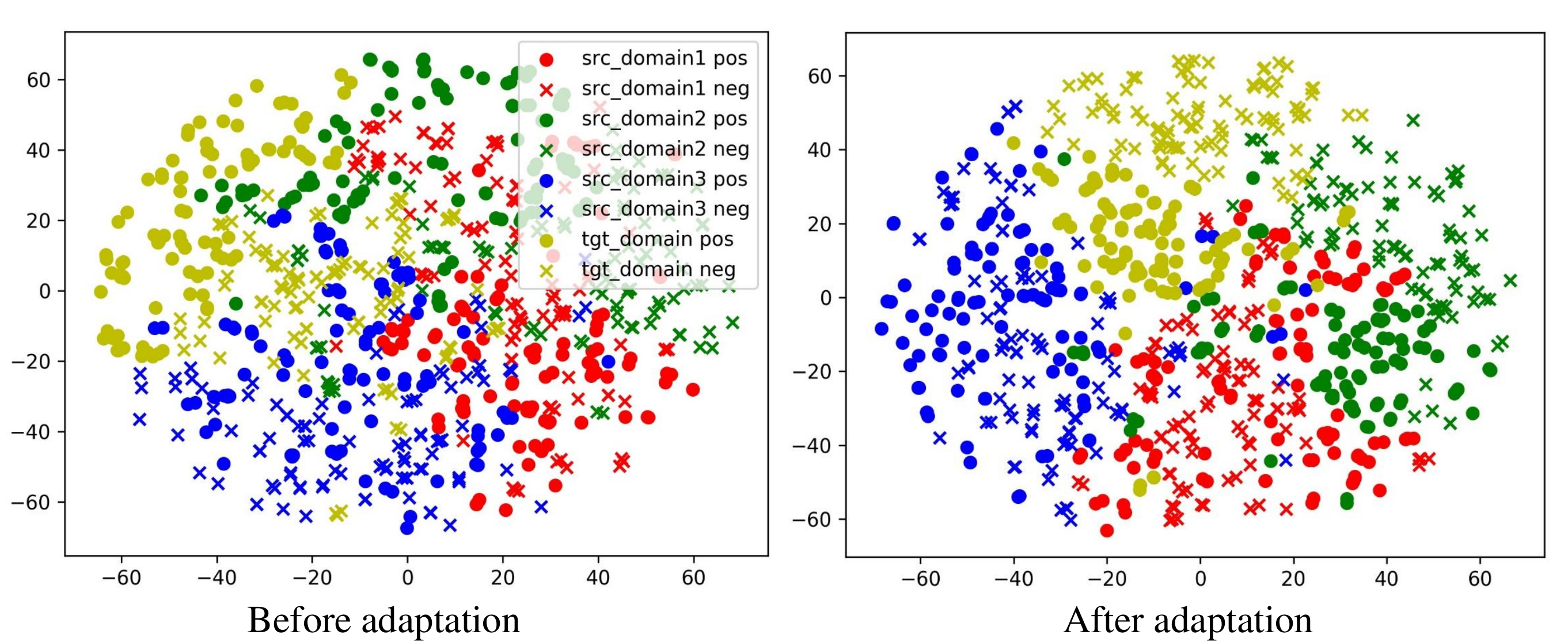}
\caption{The t-SNE visualizations of our model with adaptation (right) and without adaptation (left).}
\label{fig:fea}
\end{figure}
\subsection{Visualizations of the Proposed Method}
Considering that O\&M\&I to C set has the most significant domain shift, we visualize the feature distribution learned by Ours\_wo/adapt (before adaptation) and Ours (after adaptation) to analyse the influence of optimizing the adaptor at inference. As shown in Fig. \ref{fig:fea}, we randomly select $200$
samples of each category from four datasets and plot the t-SNE \cite{2008Visualizing} visualizations. It can be seen that after adaptation the features of fake faces and real faces on the target domain C are more compact and depart from each other further away compared to those before adaptation. It also can be find that the features of some source domains (e.g. src\_domain1) are more inseparable. However, it is worth noting that at inference we only need to make the features more suitable for the test domain to get higher testing performance. If we want to make good predictions on the source domains as before, we can remove the adaptor, since only parameters of the adaptor are updated.

\section{Conclusion}
In this work, we propose a novel self-domain adaptation framework to leverage the information of the test domain to improve the performance of spoofing face detection at inference. To achieve this goal, a meta-learning based adaptor learning algorithm is proposed for better adaptor initialization. Besides, an effective adaptor loss is proposed to make it possible to update the adaptor unsupervisedly. Extensive experiments show that our method is effective and achieves promising results on four public datasets. Moreover, we open up a new direction for face anti-spoofing to extract discriminative features from domain-specific information of the test domain to further boost the performance.

\section{Acknowledgments}
This work is supported by the National Key Research and Development Project of China (2018YFC0807702).

\bibliographystyle{aaai}
\bibliography{bibfile}

\end{document}